\documentclass[runningheads]{llncs}

\usepackage{eccv}

\usepackage{eccvabbrv}

\usepackage{graphicx}
\usepackage{booktabs}
\usepackage{multicol,multirow}
\usepackage{amssymb}
\usepackage{comment}
\usepackage[dvipsnames]{xcolor}
\definecolor{deltaup}{RGB}{0,120,60}
\definecolor{deltadn}{RGB}{180,30,30}

\usepackage[accsupp]{axessibility}  

\usepackage{orcidlink}

\begin{document}

\title{Bounding-Box Trajectories Matter \\ for Video Anomaly Detection} 


\author{
Inpyo Song\orcidlink{0009-0005-1281-9149} \and
Jangwon Lee\orcidlink{0000-0002-6601-7302}
}

\authorrunning{I.~Song and J.~Lee}

\institute{
Sungkyunkwan University,
Seoul, Republic of Korea\\
\email{\{songinpyo,leejang\}@skku.edu}
}

\maketitle

\begin{abstract}
Video anomaly detection is critical for public safety and security, yet remains highly challenging despite extensive research due to large variations in appearance, viewpoint, and scene dynamics. 
Among existing approaches, human pose-based methods have emerged as a major line of research, showing strong performance since many anomalies in public datasets involve humans and pose representations are robust to appearance changes while providing compact motion descriptions.
However, these methods often overlook bounding-box trajectories, although such information is inherently available in pose-based pipelines. In this paper, we explicitly leverage these trajectories as a primary anomaly cue. 
We present TrajVAD, a framework that models multi-class bounding-box trajectories using normalizing flows to learn normal kinematic patterns.
Its trajectory-only variant, TrajVAD-T, eliminates pose estimation, reaches 87.7 AP on ShanghaiTech, and achieves the best results on MSAD among compared methods.
TrajVAD-P adds a reliability-gated pose branch and improves performance to 88.6 AUROC and 90.9 AP on ShanghaiTech, establishing bounding-box trajectories as an effective modality for video anomaly detection.
\end{abstract}

\section{Introduction}
\label{sec:intro}

\begin{figure}[t]
    \centering
    \includegraphics[width=\linewidth]{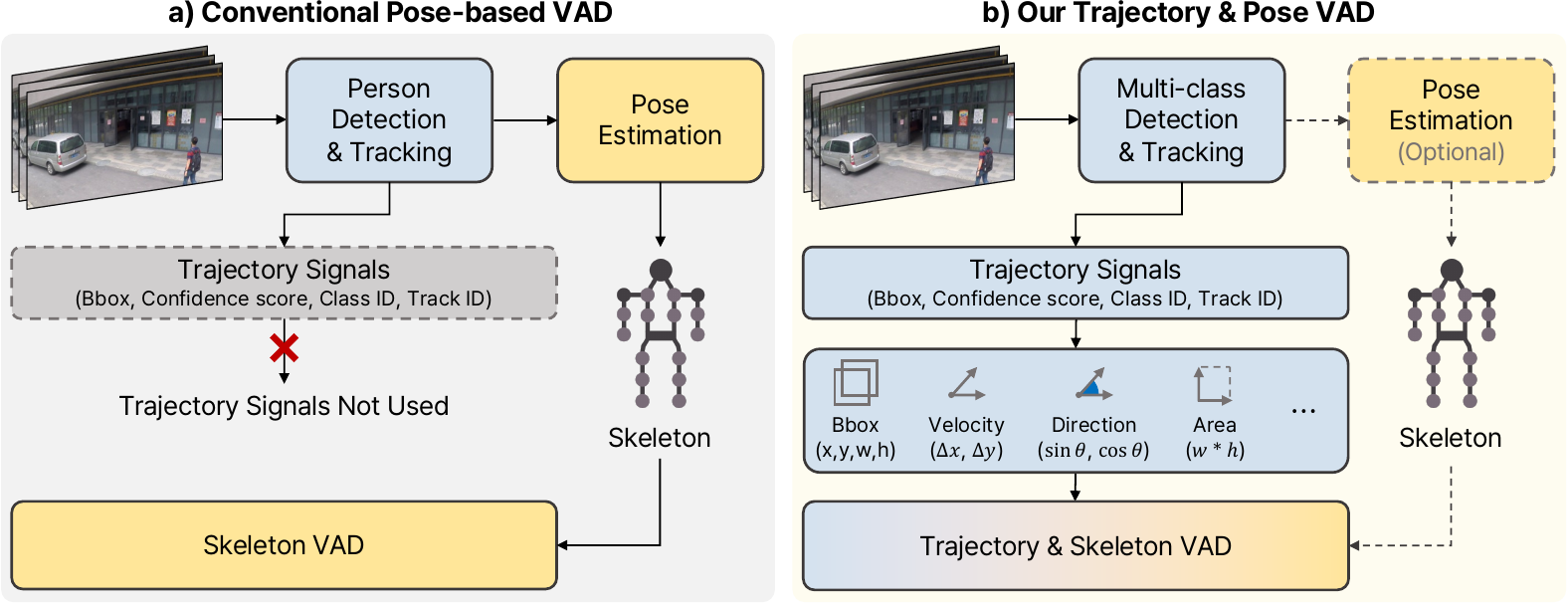}
    \caption{
        Pose-based VAD methods (left) score anomalies from skeleton sequences and are limited to person-class tracks.
        TrajVAD (right) treats multi-class bounding-box trajectories as the primary signal, applicable to any detected object.
        TrajVAD-P adds an optional pose branch (dashed) activated only for human tracks when pose is reliable.
    }
    \label{fig:teaser}
\end{figure}

Video anomaly detection (VAD) aims to discover events in videos that deviate from normal behavioral patterns. 
It serves as a key capability in intelligent surveillance and traffic monitoring systems, where a wide variety of dynamic scenes must be interpreted automatically~\cite{view360,song2024video,rare}. 
Despite decades of progress, VAD remains challenging because scenes often contain diverse objects, including pedestrians, cyclists, and vehicles, and anomalous events can involve any of them~\cite{ffp,msad}. 
Moreover, practical deployment often demands real-time performance, 
as delayed detection offers limited utility in safety-critical environments. 
A practical VAD system must therefore generalize across multiple object classes while maintaining computational efficiency for real-time or large-scale operation.

Early deep learning approaches addressed this problem through pixel-level reconstruction or prediction~\cite{conv-ae,ffp,hf2,jigsaw,fpdm,mulde,gcl}. 
While these methods capture global scene context, they inherently entangle task-relevant motion signals with extraneous appearance factors. 
Consequently, variations in lighting, background texture, or privacy-sensitive content often dominate the loss landscape, limiting the model's ability to focus purely on behavioral anomalies. 
To mitigate this sensitivity and decouple motion from appearance, pose-based methods have emerged as a popular alternative, modeling skeleton keypoint sequences to detect anomalies~\cite{stg-nf,seeker,mocodad}. 
However, while effective in person-dominated scenes, these methods remain inherently human-centric, producing no anomaly cues for non-human entities such as vehicles and degrading significantly when keypoint estimation becomes unreliable under far-field views or heavy occlusion.

To address these limitations, we turn our attention to the bounding-box trajectory, an intermediate signal often overlooked in these frameworks. 
We observe that every pose-based pipeline begins with detection and tracking, producing per-instance bounding-box trajectories before skeleton extraction even starts. 
These trajectories thus naturally exist for every detected object class, not only persons, yet existing pipelines treat them merely as bookkeeping for identity association and continue to score anomalies solely from keypoints (Figure~\ref{fig:teaser}, left). 
Building on this observation, we revisit earlier attempts at trajectory-based anomaly detection. 
Although trajectory analysis has a history in VAD, 
prior works have either modeled only center-point paths~\cite{piciarelli2008trajectory} or coupled trajectories with skeleton predictions, 
thereby inheriting the person-specific constraints of pose methods~\cite{mped-rnn,mtp,tsgad,traj-rec}. 
Consequently, the potential of multi-class bounding-box trajectories as a primary, standalone anomaly modality remains largely unexplored on modern VAD benchmarks.

To bridge this gap, we propose TrajVAD, a new framework that treats multi-class bounding-box trajectories as the primary anomaly signal and models their distribution through a likelihood-based objective trained solely on normal data (Figure~\ref{fig:teaser}, right). 
In the proposed framework, each tracked instance is represented as a sequence of trajectory-derived features, and a normalizing flow scores anomalies via the negative log-likelihood under the learned normal distribution.
TrajVAD features two variants that share the same trajectory backbone. TrajVAD-T operates purely on bounding-box trajectories and requires no pose estimation, while TrajVAD-P extends it with a reliability-gated pose branch activated only for human tracks when pose estimates are sufficiently confident.

We comprehensively evaluate the proposed TrajVAD framework on three widely used video anomaly detection benchmarks, ShanghaiTech, UBnormal, and MSAD, to validate its generality across diverse environments.
On ShanghaiTech, TrajVAD-T remains competitive with pose-based baselines without pose estimation, showing that trajectory kinematics alone carry strong anomaly cues.
TrajVAD-P further surpasses all compared pose-based methods on both AUROC and AP by incorporating the reliability-gated pose branch only for human tracks. 
Furthermore, leveraging multi-class trajectory coverage yields the largest gains over pose-based methods on MSAD, where non-human anomalies are prevalent.

Therefore, this work makes three key contributions:
\begin{itemize}
    \renewcommand{\labelitemi}{$\bullet$}
    \item We establish multi-class bounding-box trajectories as a competitive primary modality for video anomaly detection, covering all detectable object classes without requiring pose estimation.
    \item We show that trajectory features, a free byproduct of the detection and tracking pipeline that pose-based methods already execute, can eliminate the pose estimation stage and its associated computational cost.
    \item We evaluate TrajVAD-T and TrajVAD-P on ShanghaiTech, UBnormal, and MSAD, where both variants run faster than pose-based methods and achieve the highest scores among all compared methods on ShanghaiTech and MSAD.
\end{itemize}

\section{Related Work}
\label{sec:related}

\subsection{Appearance-based Video Anomaly Detection}
A major line of video anomaly detection models visual normality from RGB frames, optical flow, or deep visual features.
Reconstruction-based methods learn to reproduce normal frames or latent visual representations with convolutional, recurrent, or memory-augmented autoencoders, and use reconstruction discrepancies as anomaly evidence~\cite{conv-ae,srnn,lsa,gong2019memorizing}.
Prediction-based methods learn temporal regularity by forecasting future frames, completing missing video events, or coupling optical-flow reconstruction with flow-guided frame prediction~\cite{ffp,vec,hf2}.
Recent approaches shift the scoring space from raw pixels to learned representations, using diffusion-based feature prediction, generative cooperative learning, or multiscale density estimation over visual features~\cite{fpdm,gcl,mulde}.
These frame-centric models preserve global scene context and can provide spatial evidence without requiring explicit detection or tracking.
However, their scores are not naturally associated with persistent object instances, making object-wise temporal cues such as motion, scale variation, and changes in spatial extent less directly accessible.

\subsection{Object-centric and Pose-based Video Anomaly Detection}

A complementary paradigm, broadly termed object-centric approaches, grounds anomaly evidence in detected instances by attaching normality models to 
bounding box crops or tracklets~\cite{hinami2017joint,ssmtl++,wu2022explainable,singh2024tracklet,doshi2023towards,song2026instance} or leveraging scene-graph relations~\cite{singh2023eval,fusedvision}.
These pipelines rely on detection and tracking to isolate objects, yet treat the resulting bounding-box trajectories merely as bookkeeping for identity association, overlooking the kinematic information they encode—translation, scale variation, and aspect-ratio dynamics.

Pose-based methods, a specialized subset, model each person as a skeleton keypoint sequence~\cite{mped-rnn,gepc,normal-graph,gcae-lstm,posecvae,coskad,mocodad,stg-nf,seeker,wu2025flow}, with architectures progressing from recurrent prediction~\cite{mped-rnn} and graph clustering~\cite{gepc} through spatial-temporal graph convolutions~\cite{normal-graph,gcae-lstm} and normalizing flows~\cite{stg-nf} to diffusion models~\cite{mocodad,karami2025graph}.
While skeletons are compact and background-invariant, they are inherently human-centric, provide no signal for non-human objects, and become unreliable under occlusion or far-field views.
Most critically, every pose-based pipeline already runs detection and tracking as a prerequisite, so per-instance bounding-box trajectories exist as a virtually \textbf{free} intermediate product, yet current methods overlook this signal in favor of the computationally expensive pose estimation stage.

\subsection{Trajectory-based Video Anomaly Detection}
Trajectory-level anomaly detection has a long history.
Classical work models the distribution of center-point paths to flag routes that deviate from learned norms~\cite{piciarelli2008trajectory}.
More recent methods extend center-point paths with multi-time scale trajectory prediction~\cite{mtp} and bi-directional trajectory prediction under pose constraints~\cite{bipoco}.
TrajREC~\cite{traj-rec} further introduces holistic multitask trajectory modeling by jointly reconstructing past, present, and future segments with a contrastive objective.
TSGAD~\cite{tsgad} includes a trajectory branch that predicts person-center coordinates obtained from the detector bounding box alongside a pose branch.
However, in practice, these trajectory components remain closely tied to human-centric settings and person-level paths.
Moreover, existing approaches like TrajREC and TSGAD do not extend their analysis to multi-class bounding-box kinematics, 
thereby neglecting non-person tracks and valuable dynamic signals such as scale evolution and detector confidence. 
Consequently, the potential of utilizing generic bounding-box trajectories from standard object trackers, which naturally cover all detected object categories, as a primary anomaly modality remains unexplored in video anomaly detection.
TrajVAD addresses this gap by establishing trajectory-derived features as the core, class-agnostic input, while reserving pose estimation strictly as an optional refinement for human tracks.

\section{Method}
\label{sec:method}

\begin{figure}[t]
    \centering
    \includegraphics[width=\linewidth]{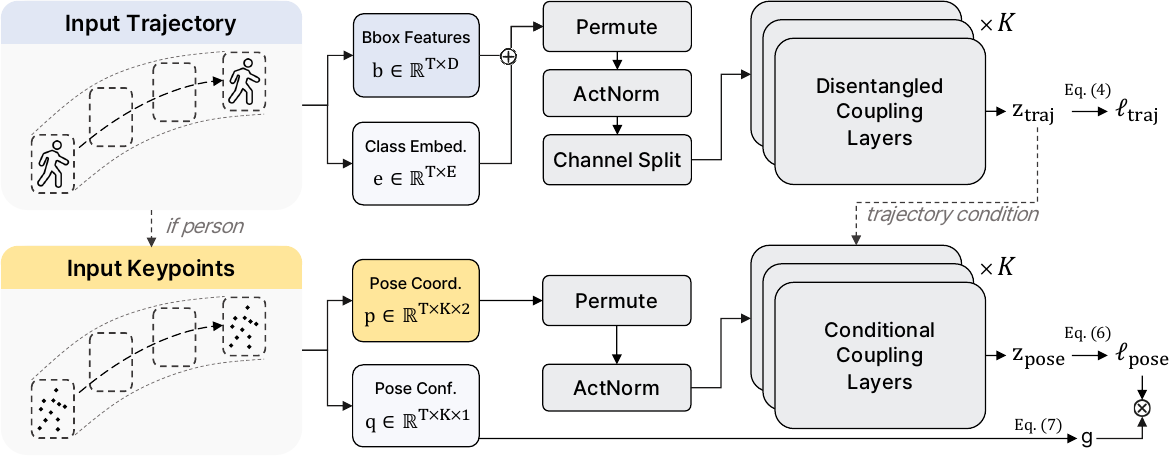}
    \caption{
        TrajVAD pipeline.
        Multi-class tracks from detection and tracking are encoded as standardized trajectory-derived feature sequences and conditioned on class embeddings.
        TrajVAD-T (top row) maps segments through a normalizing flow and uses the negative log-likelihood as the anomaly score.
        TrajVAD-P (both rows) adds a pose branch conditioned on the trajectory latent $\mathbf{z}_\text{traj}$, gated by pose reliability $g$.
    }
    \label{fig:pipeline}
\end{figure}

\subsection{Overview}
We present a framework that captures normal motion patterns directly from multi-class bounding-box trajectories.
Given an input video, the pipeline first employs an object detector and a multi-class tracker to generate per-instance tracks. 
These tracks are encoded as sequences of trajectory-derived features, augmented with learnable class embeddings to capture category-specific dynamics. 
To quantify normality, a normalizing flow model learns the distribution of these features and assigns a likelihood score to each track segment.

We propose two variants: TrajVAD-T, which relies solely on trajectory data, and TrajVAD-P, which extends the former with a pose branch to refine human tracks when skeleton estimates are available.
Both variants share a unified trajectory backbone and scoring mechanism, as illustrated in Figure~\ref{fig:pipeline}.

\begin{table}[t]
    \centering
    \caption{
        Bounding-box trajectory features (27 total).
        Coordinates and dimensions are normalized to $[0,1]$ by frame resolution before feature computation.
    }
    \label{tab:features}
    \small
    \begin{tabular}{@{}l p{6.0cm}@{}}
        \toprule
        Group & Features \\
        \midrule
        State (6) &
            Center ($c_x$, $c_y$),
            box size (w, h, area, ratio) \\
        \midrule
        Temporal dynamics (10) &
            Velocity ($v_x$, $v_y$, speed),
            acceleration ($a_x$, $a_y$, acc),
            direction ($\sin\theta$, $\cos\theta$),
            jerk, curvature \\
        \midrule
        Geometric dynamics (2) &
            Box expansion rate, aspect-ratio velocity \\
        \midrule
        Pseudo-physical (2) &
            Kinetic energy proxy (area $\times$ speed$^2$),
            path efficiency \\
        \midrule
        Perspective-normalized (6) &
            Speed per box dim ($v/h$, $v/w$, $v/\text{area}$),
            acceleration per box dim ($a/h$, $a/w$, $a/\text{area}$) \\
        \midrule
        Confidence (1) & Detector confidence (conf) \\
        \bottomrule
    \end{tabular}
\end{table}

\subsection{Trajectory Representation}

\subsubsection{Feature extraction.}
To model motion dynamics effectively, we begin with the bounding-box sequences produced by the tracker.
Each box encodes both spatial position and size at every frame but, in its raw form, does not yet capture temporal dependencies.
We normalize all box coordinates, including center position ($c_x$, $c_y$) and dimensions ($w$, $h$), to $[0,1]$ by dividing by frame resolution, making features resolution-independent.
From these normalized quantities, we construct a feature vector $\mathbf{b}_t \in \mathbb{R}^D$ at each frame.
The features are organized into six groups (Table~\ref{tab:features}).
State features record instantaneous position and box geometry.
Temporal dynamics capture velocity, acceleration, and higher-order derivatives of center motion.
Geometric dynamics track how box shape changes over time.
Pseudo-physical features combine size and speed into energy and efficiency proxies.
Perspective-normalized features divide translational quantities by box dimensions to reduce dependence on camera-to-subject distance.
Finally, detector confidence provides a scalar signal that correlates with occlusion and unusual appearance.
All features are standardized using training-split statistics.
Exact definitions for each feature are provided in the supplementary material.
To ensure stable measurements of motion, the bounding-box coordinates from the tracker are smoothed before feature extraction.
Each track is then segmented into overlapping windows of $T$ frames, and segments shorter than $T$ are discarded.

\subsubsection{Class embedding.}
Each trajectory segment is further contextualized by its associated object class.
Because the trajectory features defined above rely exclusively on generic bounding-box statistics, they naturally generalize across all detected object categories, including but not limited to the pedestrian and person classes.
We associate each segment with a class label $c \in \{0, \ldots, C{-}1\}$.
We then retrieve a trainable embedding $\mathbf{e} \in \mathbb{R}^E$ for $c$ and concatenate it to the feature vector at every time step, yielding
\begin{equation}
    \mathbf{x}_t = [\mathbf{b}_t \;\|\; \mathbf{e}] \in \mathbb{R}^{D+E}.
\label{eq:class_emb}
\end{equation}
This allows a single shared flow to distinguish class-specific normality patterns.

\subsection{TrajVAD-T: Trajectory Normalizing Flow}

TrajVAD-T models the distribution of normal trajectory segments with a normalizing flow.
The flow maps a segment $\mathbf{x} \in \mathbb{R}^{T \times (D+E)}$ to a latent vector $\mathbf{z}_\text{traj}$ through a sequence of invertible transforms and evaluates log-likelihood under a Gaussian prior.

The first transform is ActNorm~\cite{glow}, a data-adaptive affine layer initialized from the first training batch:
\begin{equation}
    \mathbf{z}_0 = (\mathbf{x} + \boldsymbol{\beta}) \odot \boldsymbol{\sigma}
\label{eq:actnorm}
\end{equation}
where $\boldsymbol{\beta}$ and $\boldsymbol{\sigma}$ are learned channel-wise bias and scale parameters.
The normalized representation $\mathbf{z}_0$ then passes through $K$ alternating affine coupling layers~\cite{dinh2016density}.
Each layer splits the feature channels into two halves $(\mathbf{u}, \mathbf{v})$ and transforms one half while keeping the other fixed:
\begin{equation}
    \mathbf{u}' = \mathbf{u}, \qquad
    \mathbf{v}' = (\mathbf{v} + t(\mathbf{u})) \odot \mathbf{a}(\mathbf{u}), \quad \mathbf{a}(\mathbf{u})=\operatorname{sigmoid}(s(\mathbf{u})+2)+\epsilon
\label{eq:coupling}
\end{equation}
where a causal convolutional subnetwork predicts the translation $t(\cdot)$ and raw scale $s(\cdot)$ parameters.
The subnet uses causal 1-D convolutions over the temporal axis with increasing dilation to expand the receptive field within the segment.
Alternating layers swap which half is transformed, ensuring all channels interact across the stack.
The final output latent is $\mathbf{z}_\text{traj} = f_K \circ \cdots \circ f_1(\mathbf{z}_0)$.

The unnormalized log-likelihood of a trajectory segment is
\begin{equation}
    \ell_\text{traj} = \log p(\mathbf{z}_\text{traj}) + \log\bigl|\det J\bigr|
\label{eq:loglik_traj}
\end{equation}
where $p(\mathbf{z}_\text{traj}) = \mathcal{N}(\mathbf{z}_\text{traj};\,\mu_0 \mathbf{1},\,\mathbf{I})$ with a fixed offset $\mu_0$, and $\log|\det J|$ accumulates log-determinants from ActNorm and all coupling layers.
The training objective minimizes the segment-normalized negative log-likelihood:
\begin{equation}
    \mathcal{L}_\text{traj} = -\frac{\ell_\text{traj}}{T(D{+}E)}.
\label{eq:loss_traj}
\end{equation}
At test time, $\mathcal{L}_\text{traj}$ serves as the per-segment anomaly score.

\subsection{TrajVAD-P: Reliability-Gated Pose Branch}

TrajVAD-P extends TrajVAD-T with an optional pose branch for person-class segments that have a valid pose estimate.
We extract 17-keypoint poses using AlphaPose~\cite{alphapose} and split each estimate into keypoint coordinates $\mathbf{p} \in \mathbb{R}^{T \times 2J}$ and per-keypoint confidence scores $\mathbf{q} \in \mathbb{R}^{T \times J}$ ($J{=}17$), both aligned to the same track window.

The pose branch is a separate normalizing flow with its own ActNorm and $K_p$ coupling layers.
Each pose coupling layer conditions its scale and translation subnets on the trajectory latent $\mathbf{z}_\text{traj}$.
The pose flow therefore models pose normality conditioned on the trajectory cues.
The pose log-likelihood is computed analogously to the trajectory branch:
\begin{equation}
    \ell_\text{pose} = \log p(\mathbf{z}_\text{pose}) + \log\bigl|\det J_\text{pose}\bigr|.
\label{eq:loglik_pose}
\end{equation}
 
The pose branch activates only when the segment belongs to the person class ($g_\text{cls}{=}1$) and a valid pose estimate exists ($g_\text{valid}{=}1$).
To prevent unreliable pose from corrupting the trajectory score, the pose contribution is further scaled by the mean keypoint confidence $\bar{q} \in [0,1]$.
These three conditions are combined into a single scalar gate:
\begin{equation}
    g = g_\text{cls} \cdot g_\text{valid} \cdot \bar{q}.
\label{eq:gate}
\end{equation}
When $g = 0$, TrajVAD-P reduces exactly to TrajVAD-T.

The combined score normalizes the joint log-likelihood by the effective dimensionality:
\begin{equation}
    \mathcal{L}_\text{total} = -\frac{\ell_\text{traj} + \lambda\, g\, \ell_\text{pose}}{T \cdot (D_\text{traj} + D_\text{pose} \cdot g)}
\label{eq:loss_total}
\end{equation}
where $\ell_\text{traj} = \log p(\mathbf{z}_\text{traj}) + \log|\det J_\text{traj}|$ and $\ell_\text{pose}$ are the unnormalized log-likelihoods of the trajectory and pose flows respectively, $\lambda$ is a pose weight hyperparameter, and $D_\text{traj}$, $D_\text{pose}$ are the trajectory and pose feature dimensions.
The denominator $T \cdot (D_\text{traj} + D_\text{pose} \cdot g)$ normalizes by the effective dimensionality, so that the score scale remains consistent whether pose is active or not.
When pose is inactive ($g = 0$), this reduces to $\mathcal{L}_\text{traj}$ (Eq.~\ref{eq:loss_traj}).

\section{Experimental Results}
\label{sec:experimental_results}

\subsection{Datasets}

We evaluate the proposed approach on three public VAD benchmarks spanning real and synthetic surveillance scenarios.

\textbf{ShanghaiTech}~\cite{ffp} contains 330 normal training videos and 107 test videos recorded by 13 fixed cameras on a university campus.
Abnormal events include running, fighting, skateboarding, and cycling, and 6 test videos feature non-human anomalies (e.g., vehicles entering pedestrian areas).

\textbf{UBnormal}~\cite{ubnormal} is a synthetic benchmark rendered with Cinema4D across 29 scenes, containing 268 training, 64 validation, and 211 test videos with both frame- and pixel-level annotations.
It includes normal actions such as walking, talking, and sitting, and staged anomalies such as falling, lying down, stealing, and dancing.

\textbf{MSAD}~\cite{msad} is the most diverse benchmark considered here, with 720 videos covering 55 abnormal event types across varied locations and viewpoints.
It contains many non-human anomalies (e.g., industrial and traffic events) that yield no skeleton keypoints.
We additionally consider its Human-Related subset (MSAD-HR), which includes seven categories (e.g., Fighting, Robbery, and Traffic Accident) and often involves non-person objects alongside humans, making it our main benchmark for multi-class trajectory coverage.

\textbf{HR subsets.}
Since skeleton-based methods provide no anomaly signal for non-human events, we follow prior protocols~\cite{mped-rnn,stg-nf,seeker,msad} and define Human-Related (HR) subsets for all three datasets.

\subsection{Evaluation Metrics}

We assign an anomaly score to each frame in the test set, concatenate scores across all test videos, and compute the Area Under the Receiver Operating Characteristic curve (AUROC) and Average Precision (AP) against the binary ground-truth labels.
This micro-averaged protocol follows prior work~\cite{stg-nf,seeker}.

\subsection{Implementation Details}
\label{sec:implementation_details}
We use YOLOX~\cite{yolox} for detection and OSNet~\cite{osnet} for re-identification, 
following STG-NF~\cite{stg-nf} and SeeKer~\cite{seeker}.
Unlike these pose-based baselines, we replace the pose-specific tracker with ByteTrack~\cite{zhang2022bytetrack} 
to support multi-class tracking beyond persons.
Tracks are segmented into windows of $T{=}12, 16, 24$ for UBnormal, ShanghaiTech, and MSAD, respectively.
Before windowing, bounding-box coordinates are smoothed with a Kalman filter.
Missing detections spanning at most 10 frames are filled by linear interpolation, whereas longer gaps split the track.
This preprocessing handles transient detector dropouts without forcing long occluded intervals into a single trajectory.
The trajectory flow uses $K{=}6$ coupling layers for TrajVAD-T and $K{=}18$ for TrajVAD-P, 
with a 3-dimensional class embedding over the 80 COCO classes.
The Gaussian prior uses a mean offset $\mu_0{=}3$.
Further implementation details are provided in the supplementary material.

\begin{table*}[t]
    \centering
    \caption{
        Comparison with RGB, flow, multimodal, pose-based, and trajectory-based methods on ShanghaiTech~\cite{ffp}.
        SHT-HR denotes the human-related subset.
        The supplementary material reports the property breakdown and localization metrics.
    }
    \label{tab:shanghaitech}
    \begingroup
    \setlength{\tabcolsep}{6pt}
    \renewcommand{\arraystretch}{1.03}
    \scriptsize
    \begin{tabular}{@{}l l cccc @{} }
        \toprule
        \multirow{2}{*}{Method} & \multirow{2}{*}{Venue}
        & \multicolumn{2}{c}{AUROC}
        & \multicolumn{2}{c}{AP} \\
        \cmidrule(lr){3-4} \cmidrule(lr){5-6}
         & & SHT & SHT-HR & SHT & SHT-HR \\
        \midrule
        \multicolumn{6}{@{}l}{\textit{RGB, flow, and multimodal baselines}} \\
        sRNN~\cite{srnn}            & ICCV'17   & 68.0 & --   & --   & -- \\
        Conv-AE~\cite{conv-ae}      & CVPR'16   & 70.4 & 69.8 & --   & -- \\
        LSA~\cite{lsa}              & CVPR'19   & 72.5 & --   & --   & -- \\
        FFP~\cite{ffp}              & CVPR'18   & 72.8 & 72.7 & --   & -- \\
        VEC~\cite{vec}              & ACM MM'20 & 74.8 & --   & --   & -- \\
        HF$^2$~\cite{hf2}           & ICCV'21   & 76.2 & --   & --   & -- \\
        FPDM~\cite{fpdm}            & ICCV'23   & 78.6 & --   & --   & -- \\
        CAE-SVM~\cite{cae-svm}      & ICCV'17   & 78.7 & --   & --   & -- \\
        GCL~\cite{gcl}              & CVPR'22   & 79.6 & --   & --   & -- \\
        BA-AED~\cite{ba-aed}        & TPAMI'22  & 82.7 & --   & --   & -- \\
        SSMTL++~\cite{ssmtl++}      & CVIU'23   & 83.8 & --   & --   & -- \\
        Jigsaw~\cite{jigsaw}        & ECCV'22   & 84.2 & 84.7 & --   & -- \\
        MULDE~\cite{mulde}          & CVPR'24   & \underline{86.7} & -- & -- & -- \\
        \midrule
        \multicolumn{6}{@{}l}{\textit{Pose and trajectory baselines}} \\
        BiPOCO~\cite{bipoco}        & arXiv'22  & --   & 74.9 & --   & -- \\
        MPED-RNN~\cite{mped-rnn}    & CVPR'19   & 73.4 & 75.4 & --   & -- \\
        MTP~\cite{mtp}              & WACV'20   & 76.0 & 77.0 & --   & -- \\
        GEPC~\cite{gepc}            & CVPR'20   & 76.1 & 74.8 & --   & -- \\
        PoseCVAE~\cite{posecvae}    & ICPR'21   & --   & 75.5 & --   & -- \\
        Normal Graph~\cite{normal-graph}
                                    & NC'21     & --   & 76.5 & --   & -- \\
        COSKAD~\cite{coskad}        & PR'24     & --   & 77.1 & --   & -- \\
        GCAE-LSTM~\cite{gcae-lstm}  & NC'22     & --   & 77.2 & --   & -- \\
        MoCoDAD~\cite{mocodad}      & ICCV'23   & --   & 77.6 & --   & -- \\
        TrajREC~\cite{traj-rec}     & WACV'24   & --   & 77.9 & --   & -- \\
        TSGAD-Traj~\cite{tsgad}     & WACV'24   & 67.8 & 68.4 & 61.2 & 61.3 \\
        TSGAD~\cite{tsgad}          & WACV'24   & 80.6 & 81.5 & 73.9 & 74.2 \\
        GiCiSAD~\cite{karami2025graph}
                                    & WACV'25   & --   & 78.0 & --   & -- \\
        STG-NF~\cite{stg-nf}        & ICCV'23   & 85.9 & \underline{87.4} & 77.6 & 81.4 \\
        SeeKer~\cite{seeker}        & ICCV'25   & 85.5 & 86.9 & 80.0 & 81.5 \\
        \textbf{TrajVAD-T}          & --        & 84.9 & 84.1 & \underline{87.7} & \underline{87.5} \\
        \textbf{TrajVAD-P}          & --        & \textbf{88.6} & \textbf{88.6} & \textbf{90.9} & \textbf{91.1} \\
        \bottomrule
    \end{tabular}
    \endgroup
\end{table*}

\subsection{Comparison with State-of-the-Art Methods}

\subsubsection{ShanghaiTech~\cite{ffp}.}

Table~\ref{tab:shanghaitech} reports results on ShanghaiTech.
ShanghaiTech anomalies are predominantly motion-defined events such as running, fighting, and skateboarding in pedestrian areas, where translational dynamics and scale changes in bounding boxes provide direct discriminative signal.
TrajVAD-T trails STG-NF by 1.0 AUROC point but achieves 87.7\% AP, exceeding STG-NF and SeeKer by 10.1 and 7.7 AP points, respectively.
TrajVAD-P achieves the highest AUROC of 88.6 on both ShanghaiTech and ShanghaiTech-HR, with AP of 90.9\% and 91.1\% respectively, surpassing all compared methods on both metrics.
This gain indicates that pose cues provide complementary information for human-centric anomalies whose body configuration changes are not fully captured by bounding-box kinematics.

\begin{table*}[t]
    \centering
    \caption{
        Comparison with RGB, flow, multimodal, pose-based, and trajectory-based methods on UBnormal~\cite{ubnormal}.
        UBn-HR denotes the human-related subset.
        The supplementary material reports the property breakdown and localization metrics.
    }
    \label{tab:ubnormal}
    \begingroup
    \setlength{\tabcolsep}{6pt}
    \renewcommand{\arraystretch}{1.03}
    \scriptsize
    \begin{tabular}{@{}l l cccc @{} }
        \toprule
        \multirow{2}{*}{Method} & \multirow{2}{*}{Venue}
        & \multicolumn{2}{c}{AUROC}
        & \multicolumn{2}{c}{AP} \\
        \cmidrule(lr){3-4} \cmidrule(lr){5-6}
         & & UBn & UBn-HR & UBn & UBn-HR \\
        \midrule
        \multicolumn{6}{@{}l}{\textit{RGB, flow, and multimodal baselines}} \\
        Jigsaw~\cite{jigsaw}        & ECCV'22   & 56.4 & --   & --   & -- \\
        BA-AED~\cite{ba-aed}        & TPAMI'22  & 59.3 & --   & --   & -- \\
        SSMTL++~\cite{ssmtl++}      & CVIU'23   & 62.1 & --   & --   & -- \\
        FPDM~\cite{fpdm}            & ICCV'23   & 62.7 & --   & --   & -- \\
        MULDE~\cite{mulde}          & CVPR'24   & 72.8 & --   & --   & -- \\
        \midrule
        \multicolumn{6}{@{}l}{\textit{Pose and trajectory baselines}} \\
        BiPOCO~\cite{bipoco}        & arXiv'22  & 50.7 & 52.3 & --   & -- \\
        GEPC~\cite{gepc}            & CVPR'20   & 53.4 & 55.2 & --   & -- \\
        MPED-RNN~\cite{mped-rnn}    & CVPR'19   & 60.6 & 61.2 & --   & -- \\
        COSKAD~\cite{coskad}        & PR'24     & 65.0 & 65.5 & --   & -- \\
        TrajREC~\cite{traj-rec}     & WACV'24   & 68.0 & 68.2 & --   & -- \\
        GiCiSAD~\cite{karami2025graph}
                                    & WACV'25   & 68.6 & 68.8 & --   & -- \\
        MoCoDAD~\cite{mocodad}      & ICCV'23   & 68.3 & 68.4 & --   & -- \\
        STG-NF~\cite{stg-nf}        & ICCV'23   & 71.8 & 71.5 & 62.7 & 67.2 \\
        SeeKer~\cite{seeker}        & ICCV'25   & \textbf{77.9} & \textbf{78.9} & \textbf{80.3} & \textbf{79.8} \\
        \textbf{TrajVAD-T}          & --        & 68.0 & 68.0 & 63.2 & 63.4 \\
        \textbf{TrajVAD-P}          & --        & \underline{73.8} & \underline{73.8} & \underline{68.3} & \underline{68.4} \\
        \bottomrule
    \end{tabular}
    \endgroup
\end{table*}

\begin{table}[t]
    \centering
    \caption{
        Comparison with pose-based and trajectory-based methods on MSAD~\cite{msad}.
        MSAD-HR denotes the human-related subset.
    }
    \label{tab:msad}
    \begingroup
    \setlength{\tabcolsep}{5pt}
    \footnotesize
    \begin{tabular}{@{}l l cccc @{}}
        \toprule
        \multirow{2}{*}{Method} & \multirow{2}{*}{Venue}
        & \multicolumn{2}{c}{AUROC}
        & \multicolumn{2}{c}{AP} \\
        \cmidrule(lr){3-4} \cmidrule(lr){5-6}
         & & MSAD & MSAD-HR & MSAD & MSAD-HR \\
        \midrule

        MoCoDAD~\cite{mocodad}   & ICCV'23   & -- & 53.9 & -- & 51.7 \\
        STG-NF~\cite{stg-nf}    & ICCV'23   & 53.8 & 55.7 & 37.5 & 56.5 \\
        TrajREC~\cite{traj-rec}   & WACV'24   & 51.2 & 54.8 & 39.8 & 53.2 \\
        SeeKer~\cite{seeker}    & ICCV'25   & -- & 61.1 & -- & \underline{60.1} \\

        \textbf{TrajVAD-T}       & --        & \textbf{57.2} & \textbf{69.7} & \textbf{42.7} & \textbf{60.4} \\
        \textbf{TrajVAD-P}       & --        & \underline{55.5} & \underline{68.5} & \underline{41.5} & 58.5 \\
        \bottomrule
    \end{tabular}
    \endgroup
\end{table}

\subsubsection{UBnormal~\cite{ubnormal}.}

UBnormal presents a harder setting for trajectory-based detection (Table~\ref{tab:ubnormal}).
TrajVAD-P reaches 73.8 AUROC and 68.3 AP on UBnormal, surpassing STG-NF and all other compared pose-based methods except SeeKer, but trailing SeeKer by 4.1 AUROC points.
TrajVAD-T achieves competitive AUROC of 68.0 without pose estimation.

This gap reflects a challenge specific to UBnormal's synthetic domain.
We observe that rendered sequences produce unstable object detection, with class labels flipping between frames and confidence scores that are noisy and low.
The feature ablation in Table~\ref{tab:feature_ablation} shows that detector confidence is the most impactful single feature for TrajVAD-T ($-2.4$ AUROC when removed), so noisy confidence on synthetic imagery directly undermines the trajectory representation.
Adding pose in TrajVAD-P recovers 5.8 AUROC points over TrajVAD-T, yet cannot fully compensate for the weakened trajectory base.
A failure case analysis illustrating these detection instabilities is provided in the supplementary material.

\subsubsection{MSAD~\cite{msad}.}

MSAD tests whether multi-class trajectory coverage translates to measurable gains over pose-only methods (Table~\ref{tab:msad}).
Unlike ShanghaiTech and UBnormal, where anomalies are predominantly motion deviations such as running or jumping, MSAD includes semantically complex categories such as robbery and vandalism, and MSAD-HR contains classes like traffic accident where the primary object is a vehicle.

On MSAD-HR, TrajVAD-T achieves 69.7 AUROC and 60.4 AP, exceeding SeeKer by 8.6 AUROC points and STG-NF by 14.0 AUROC points.
This is the largest AUROC gain among the three benchmarks.
Because pose-based models produce no signal for non-human objects, TrajVAD's uniform modeling of bounding-box kinematics across all tracked classes provides a structural advantage when non-person anomalies are prevalent.
TrajVAD-T outperforms TrajVAD-P on both MSAD-HR and MSAD, consistent with the expectation that human-pose gating provides limited benefit when the dominant anomalous objects are non-human.

On full MSAD, performance drops for all methods because categories such as Explosion and Fire produce no trackable objects; TrajVAD-T still leads with 57.2 AUROC ahead of STG-NF (53.8), but the 12.5-point gap from MSAD-HR reflects anomaly types beyond the detector's reach.

Overall, the three benchmarks highlight complementary roles of trajectory and pose cues.
Trajectory-only modeling is effective when anomalies are expressed through object-level motion or involve non-human objects, as in MSAD.
Pose augmentation is most useful for human-centric anomalies with informative body-configuration changes, as in ShanghaiTech.
UBnormal, in contrast, exposes the sensitivity of trajectory modeling to synthetic detector noise.

\begin{table}[t]
    \centering
    \caption{
        Computational cost breakdown.
        Preprocessing is per frame, and inference is per segment.
        Total includes preprocessing and inference because stride-1 sliding-window evaluation produces one new segment per frame.
    }
    \label{tab:cost}
    \begingroup
    \setlength{\tabcolsep}{5pt}
    \footnotesize
    \begin{tabular}{@{}l | cc | c | c | cc @{}}
        \toprule
        \multirow{2}{*}{Method}
        & \multicolumn{2}{c|}{Preprocess (ms/frame)}
        & Inference
        & Total
        & \multicolumn{2}{c}{SHT-HR} \\
        \cmidrule(lr){2-3} \cmidrule(lr){6-7}
        & Det+Track & +Pose & (ms/seg) & (ms/seg) & AUROC & AP \\
        \midrule

        TrajREC~\cite{traj-rec}                & 105.8 & 31.0  & 4.5    & \underline{141.3} & 77.9 & -- \\
        STG-NF~\cite{stg-nf}                & 105.8 & 31.0  & 8.8    & 145.6 & \underline{87.4} & 81.4 \\
        MoCoDAD~\cite{mocodad}               & 105.8 & 31.0  & 2,857.0  & 2,993.8 & 77.6 & -- \\
        GiCiSAD~\cite{karami2025graph}       & 105.8 & 31.0  & 1,296.1   & 1,432.9 & 78.0 & -- \\

        \midrule

        \textbf{TrajVAD-T}             & 104.4 & --  & 2.8    & \textbf{107.2} & 84.1 & \underline{87.5} \\
        \textbf{TrajVAD-P}             & 104.4 & 31.0  & 8.6    & \underline{144.0} & \textbf{88.6} & \textbf{91.1} \\
        
        \bottomrule
    \end{tabular}
    \endgroup
\end{table}

\subsection{Computational Cost}

Table~\ref{tab:cost} breaks down the end-to-end cost per temporal window (segment).
Pose-based methods use a person-specific tracker, while TrajVAD uses a general bounding-box tracker (ByteTrack~\cite{zhang2022bytetrack}).
Both trackers run in roughly 1\,ms per frame, so the detection and re-identification stages (YOLOX~\cite{yolox} + OSNet~\cite{osnet}) dominate preprocessing cost.
This explains the negligible gap between the two pipelines (105.8 vs.\ 104.4\,ms/frame).
TrajVAD-T skips pose estimation entirely, saving 31\,ms per frame compared to methods that require skeleton extraction.
MoCoDAD and GiCiSAD use diffusion-based inference, which accounts for over 90\% of their total cost.
The hardware configuration used for speed benchmarking is provided in the supplementary material.

\begin{figure}[t]
    \centering
    \includegraphics[width=0.94\linewidth]{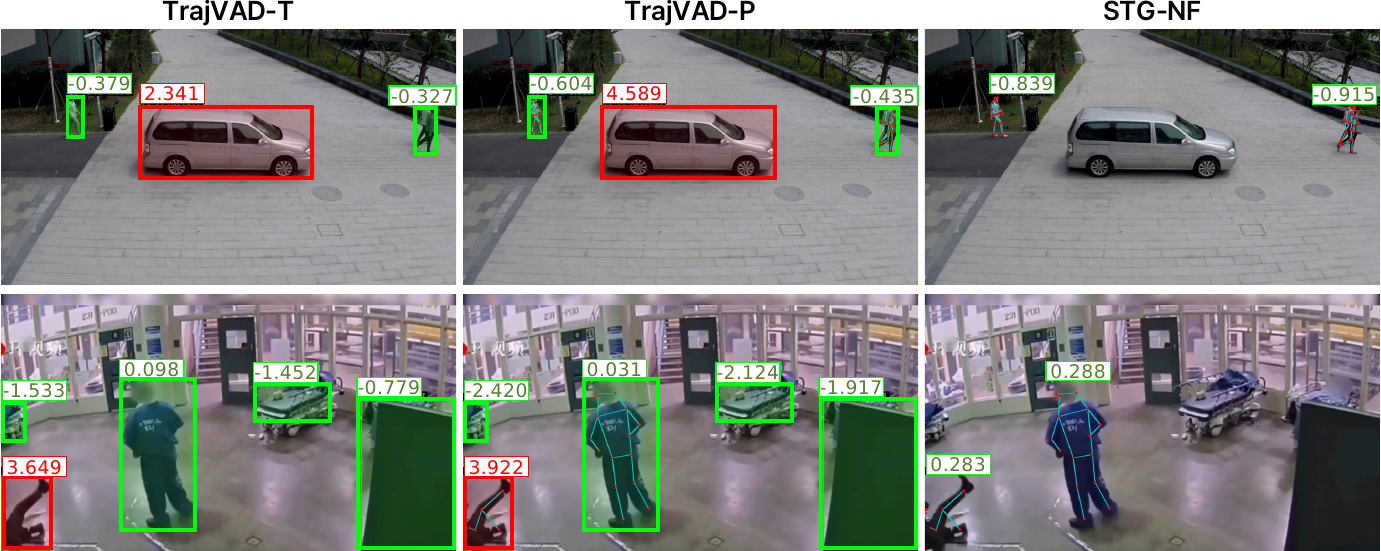}
    \caption{
        Qualitative comparison on ShanghaiTech and MSAD.
        Red boxes and anomaly scores (higher means anomaly) indicate detected anomalies.
        Top: a car in a pedestrian zone is invisible to pose-based STG-NF but flagged by TrajVAD through bounding-box kinematics.
        Bottom: partial occlusion corrupts skeleton estimation, suppressing the STG-NF anomaly score, while TrajVAD maintains detection from trajectory features.
    }
    \label{fig:qualitative}
\end{figure}

\begin{table}[t]
    \centering
    \caption{
        Feature group ablation on ShanghaiTech (leave-one-group-out, AUROC \%).
        Each row removes one feature group from the full 27-dimensional set.
        Feature groups follow Table~\ref{tab:features}.
        $D$ is the remaining feature dimensionality.
        $\Delta$ denotes AUROC change relative to the full set.
    }
    \label{tab:feature_ablation}
    \begingroup
    \setlength{\tabcolsep}{6pt}
    \footnotesize
    \begin{tabular}{@{}l | c | cc | cc @{}}
        \toprule
        \multirow{2}{*}{Features}
        & \multirow{2}{*}{$D$}
        & \multicolumn{2}{c|}{TrajVAD-T}
        & \multicolumn{2}{c}{TrajVAD-P} \\
        \cmidrule(lr){3-4} \cmidrule(lr){5-6}
        & & SHT ($\Delta$) & SHT-HR ($\Delta$) & SHT ($\Delta$) & SHT-HR ($\Delta$) \\
        \midrule
        Full 27D                     & 27 & 84.9 & 84.1 & 88.6 & 88.6 \\
        \quad$-$ State (6)           & 21 & 84.6 {\scriptsize\textbf{\color{deltadn}$-$0.3}} & 83.8 {\scriptsize\textbf{\color{deltadn}$-$0.3}} & 87.5 {\scriptsize\textbf{\color{deltadn}$-$1.1}} & 87.4 {\scriptsize\textbf{\color{deltadn}$-$1.2}} \\
        \quad$-$ Temporal (10)       & 17 & 83.9 {\scriptsize\textbf{\color{deltadn}$-$1.0}} & 83.0 {\scriptsize\textbf{\color{deltadn}$-$1.1}} & 88.2 {\scriptsize\textbf{\color{deltadn}$-$0.4}} & 88.0 {\scriptsize\textbf{\color{deltadn}$-$0.6}} \\
        \quad$-$ Geometric (2)       & 25 & 84.0 {\scriptsize\textbf{\color{deltadn}$-$0.9}} & 83.1 {\scriptsize\textbf{\color{deltadn}$-$1.0}} & 88.0 {\scriptsize\textbf{\color{deltadn}$-$0.6}} & 87.8 {\scriptsize\textbf{\color{deltadn}$-$0.8}} \\
        \quad$-$ Pseudo-physical (2) & 25 & 84.1 {\scriptsize\textbf{\color{deltadn}$-$0.8}} & 83.2 {\scriptsize\textbf{\color{deltadn}$-$0.9}} & 88.0 {\scriptsize\textbf{\color{deltadn}$-$0.6}} & 87.7 {\scriptsize\textbf{\color{deltadn}$-$0.9}} \\
        \quad$-$ Perspective (6)     & 21 & 83.4 {\scriptsize\textbf{\color{deltadn}$-$1.5}} & 82.6 {\scriptsize\textbf{\color{deltadn}$-$1.5}} & 88.6 {\scriptsize\textbf{\color{deltaup}+0.0}} & 88.6 {\scriptsize\textbf{\color{deltaup}+0.0}} \\
        \quad$-$ Confidence (1)      & 26 & 82.5 {\scriptsize\textbf{\color{deltadn}$-$2.4}} & 81.9 {\scriptsize\textbf{\color{deltadn}$-$2.2}} & 85.9 {\scriptsize\textbf{\color{deltadn}$-$2.7}} & 85.8 {\scriptsize\textbf{\color{deltadn}$-$2.8}} \\
        \bottomrule
    \end{tabular}
    \endgroup
\end{table}

\begin{table}[t]
    \centering
    \caption{
    Robustness to simulated missed detections on ShanghaiTech.
    AUROC (\%) is reported under clean detections and after randomly removing 10\% of detections per track.
    $\Delta$ denotes the change from the clean setting.
    }
    \label{tab:detection_failure}
    \begingroup
    \setlength{\tabcolsep}{5pt}
    \begin{tabular}{@{}lccc@{}}
        \toprule
        Method & Clean & 10\% removed & $\Delta$ \\
        \midrule
        TSGAD~\cite{tsgad} & 80.6 & 76.1 & {\scriptsize\color{deltadn}$-$4.5} \\
        STG-NF~\cite{stg-nf} & 85.9 & 80.7 & {\scriptsize\color{deltadn}$-$5.2} \\
        \textbf{TrajVAD-T} & 84.9 & 82.4 & {\scriptsize\textbf{\color{deltadn}$-$2.5}} \\
        \bottomrule
    \end{tabular}
    \endgroup
\end{table}

\subsection{Ablation Study}

\subsubsection{Feature group ablation.}

Table~\ref{tab:feature_ablation} evaluates each trajectory feature group through a leave-one-group-out protocol on ShanghaiTech.
Every group contributes positively for TrajVAD-T, confirming that the full 27-dimensional representation is well-utilized.

Detector confidence is the most impactful single feature for both variants ($-2.4$ ShanghaiTech AUROC for TrajVAD-T, $-2.7$ for TrajVAD-P).
Unlike all other features derived from bounding-box coordinates, 
confidence is the only signal sourced from the detector's internal state,
encoding detection quality under occlusion, truncation, and unusual appearance that no kinematic feature can recover, 
which explains why removing a single scalar causes the largest drop.
For TrajVAD-T, state features show the smallest effect ($-0.3$) because higher-order derivatives already subsume raw position and geometry.
Perspective-normalized features are the second most impactful group for TrajVAD-T ($-1.5$) but cause no degradation for TrajVAD-P ($+0.0$), as the pose branch's pelvis normalization already encodes body-relative scale, making perspective correction redundant when pose is available.

\begin{figure}[t]
    \centering
    \includegraphics[width=\linewidth]{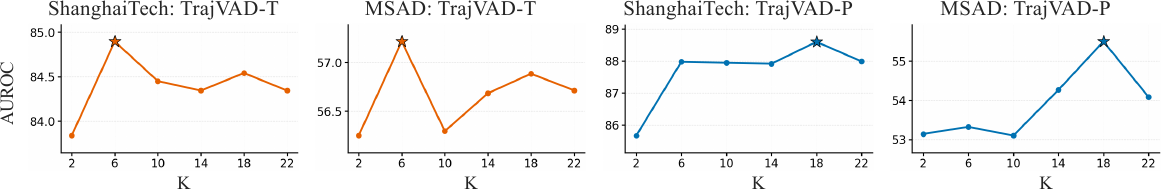}
    \caption{
        Effect of flow depth $K$ on AUROC for TrajVAD-T and TrajVAD-P on ShanghaiTech and MSAD.
        Stars mark the best $K$ per panel.
        TrajVAD-T is robust across depths, while TrajVAD-P peaks at $K{=}18$.
    }
    \label{fig:k_ablation}
\end{figure}

\subsubsection{Flow depth ablation.}

Figure~\ref{fig:k_ablation} varies the number of coupling layers $K$ from 2 to 22 on ShanghaiTech and MSAD.
TrajVAD-T is relatively insensitive to $K$ on ShanghaiTech, with AUROC ranging from 83.8 to 84.9 across the sweep and the best result at $K{=}6$.
On MSAD, TrajVAD-T shows a similar pattern, peaking at $K{=}6$ with 57.2 AUROC.
TrajVAD-P exhibits a clearer optimum at $K{=}18$ on both datasets, suggesting that the pose-conditioned flow benefits from additional depth to model the joint trajectory-pose distribution.
Based on these results, TrajVAD-T uses $K{=}6$ and TrajVAD-P uses $K{=}18$ in the main comparison tables.

\subsubsection{Robustness to detection failures.}
\label{sec:det_failure_robustness}
To evaluate robustness to upstream detection errors, we randomly remove 10\% of detections per track on ShanghaiTech and apply the preprocessing in Sec.~\ref{sec:implementation_details}.
Table~\ref{tab:detection_failure} shows that TrajVAD-T loses 2.5 AUROC points, while STG-NF and TSGAD lose 5.2 and 4.5 points, respectively.
These results indicate that bounding-box trajectory modeling remains stable under short detection gaps, especially when combined with Kalman smoothing, interpolation, and conservative track splitting.

\subsection{Qualitative Analysis}

Figure~\ref{fig:qualitative} shows two representative failure modes of pose-based detection that TrajVAD handles correctly.
In the top row, a car drives through a pedestrian area.
STG-NF assigns no anomaly score to the vehicle because pose estimation is undefined for non-person objects, whereas both TrajVAD-T and TrajVAD-P flag the car through its bounding-box trajectory alone.
In the bottom row, a person is only partially visible with legs occluded, causing skeleton estimation to fail.
STG-NF produces a low anomaly score from the corrupted pose, while TrajVAD-T still detects the anomaly from bounding-box kinematics and TrajVAD-P further amplifies the score by gating out the unreliable pose.

\section{Limitations and Future Work}
\label{sec:limitations}
One limitation of the proposed method is that TrajVAD's performance is strongly affected by the quality of its upstream detector.
As shown on UBnormal (Table~\ref{tab:ubnormal}), noisy confidence scores and class-label flipping in synthetic imagery degrade the trajectory representation.
The feature ablation in Table~\ref{tab:feature_ablation} also shows that detector confidence is the most influential feature, with a $-2.4$ AUROC drop when it is removed.
Thus, scenes with poorly calibrated confidence scores, missed detections, or unstable track IDs can reduce TrajVAD's accuracy.
Another limitation stems from the detector's class coverage.
Although the shared COCO-trained detector spans 80 categories, which is substantially broader than person-only pose pipelines, anomalies involving out-of-vocabulary objects or events that yield no foreground detection remain beyond reach.
The 12.5-point gap between MSAD-HR (69.7 AUROC) and MSAD (57.2 AUROC) quantifies this limitation.
Adopting an open-world or class-agnostic detector would extend this coverage without altering the trajectory modeling itself.

\section{Conclusion}
\label{sec:conclusion}

This paper revisited bounding-box trajectories as a primary modality for video anomaly detection.
Although these trajectories are already produced by detection and tracking pipelines, 
they are typically used only for identity association and rarely treated as anomaly evidence.
We showed that multi-class bounding-box dynamics provide strong object-level cues, enabling competitive anomaly detection without dense appearance modeling or mandatory pose estimation.
TrajVAD represents each tracked object with class-aware trajectory features and models normality with a normalizing flow.
The trajectory-only variant removes the pose estimation stage while retaining strong detection accuracy and improving efficiency over pose-dependent pipelines.
The reliability-gated pose variant further complements trajectory cues in human-centric cases, 
where body configuration provides information beyond bounding-box motion.
Experimental results on three widely used VAD benchmarks validated the proposed design. 
In particular, multi-class trajectory coverage yielded the largest gains on MSAD, whereas reliability-gated pose augmentation achieved the best performance on ShanghaiTech, demonstrating the complementary benefits of the two variants. 
These findings establish bounding-box trajectories as a simple, scalable, and previously underutilized modality for VAD.

\section*{Acknowledgements}
This work was supported by the National Research Foundation of Korea (NRF) grant (No. RS-2024-00458696), and by the NRF grant funded by the Korea government (MSIT) (No. RS-2024-00462874, Development of Sentinel AI system technology for predicting and preemptively responding to threats in open crowded environments), and by the Institute for Information \& Communications Technology Planning \& Evaluation (IITP) grants funded by the Korea government (MSIT), including the Global Scholars Invitation Program (No. RS-2024-00459638) and the Graduate School of Virtual Convergence support program (No. IITP-2026-RS-2023-00254129).

%
%
\bibliographystyle{splncs04}
\bibliography{main}

\end{document}